\documentclass[sigconf]{aamas}  % do not change this line!

\usepackage{booktabs}
\usepackage{multirow} % needed for tablesgenerator tables

\setcopyright{ifaamas}  % do not change this line!
\acmDOI{doi}  % do not change this line!
\acmISBN{}  % do not change this line!
\acmConference[ALA'20]{Adaptive Learning Agents Workshop}{May 2020}{Auckland, New Zealand}  % do not change this line!
\acmYear{2020}  % do not change this line!
\copyrightyear{2020}  % do not change this line!
\acmPrice{}  % do not change this line!

\renewcommand\footnotetextcopyrightpermission[1]{} % removes footnote with conference information in first column 
\begin{document}

\title[A Demonstration of Issues with Value-Based MORL Under Stochastic State Transitions]{A Demonstration of Issues with Value-Based Multiobjective Reinforcement Learning Under Stochastic State Transitions}  % put your title here!
%\titlenote{Produces the permission block, and copyright information}

%\subtitlenote{The full version of the author's guide is available as \texttt{acmart.pdf} document}

% AAMAS: submissions are anonymous for most tracks
%\author{Paper \#28}  % put your paper number here!

%% example of author block for camera ready version of accepted papers: don't use for anonymous submissions
%
\author{Peter Vamplew}
\orcid{0000-0002-8687-4424}
\affiliation{%
  \institution{Federation University Australia}
  \streetaddress{University Drive}
  \city{Mt Helen} 
  \state{Victoria}
  \country{Australia}
}
\email{p.vamplew@federation.edu.au}

\author{Cameron Foale}
\affiliation{%
  \institution{Federation University Australia}
  \streetaddress{University Drive}
  \city{Mt Helen} 
  \state{Victoria}
  \country{Australia}
}
\email{c.foale@federation.edu.au}

\author{Richard Dazeley}
\affiliation{%
  \institution{Deakin University}
  \streetaddress{75 Pigdons Road}
  \city{Waurn Ponds} 
  \state{Victoria}
  \country{Australia}
}
\email{r.dazeley@deakin.edu.au}

%% The example's default list of authors is too long for headers

\begin{abstract}  % put your abstract here!
We report a previously unidentified issue with model-free, value-based approaches to multiobjective reinforcement learning in the context of environments with stochastic state transitions. An example multiobjective Markov Decision Process (MOMDP) is used to demonstrate that under such conditions these approaches may be unable to discover the policy which maximises the Scalarised Expected Return, and in fact may converge to a Pareto-dominated solution. We discuss several alternative methods which may be more suitable for maximising SER in MOMDPs with stochastic transitions.
\end{abstract}

% AAMAS: the ACM CCS are not needed within AAMAS papers
%%
%% The code below should be generated by the tool at
%% http://dl.acm.org/ccs.cfm
%\begin{CCSXML}
%<ccs2012>
%  <concept_id>10010520.10010553.10010562</concept_id>
%  <concept_desc>Computer systems organization~Embedded systems</concept_desc>
%  <concept_significance>500</concept_significance>
%  <concept_id>10010520.10010575.10010755</concept_id>
%  <concept_desc>Computer systems organization~Redundancy</concept_desc>
%  <concept_significance>300</concept_significance>
%  <concept_id>10010520.10010553.10010554</concept_id>
%  <concept_desc>Computer systems organization~Robotics</concept_desc>
%  <concept_significance>100</concept_significance>
%  <concept_id>10003033.10003083.10003095</concept_id>
%  <concept_desc>Networks~Network reliability</concept_desc>
%  <concept_significance>100</concept_significance>
%</ccs2012>  
%\end{CCSXML}
%
%\ccsdesc[500]{Computer systems organization~Embedded systems}

  % put your semicolon-separated keywords here!

\maketitle

%%%%%%%%%%%%%%%%%%%%%%%%%%%%%%%%%%%%%%%%%%%%%%%%%%%%%%%%%%%%%%%%%%%%%%%%%%%%%%%%%%%%%%%%%%%%%%%%%%%%%%%%%
%% start of main body of paper

\section{Introduction}

Multiobjective reinforcement learning (MORL) aims to extend the capabilities of reinforcement learning (RL) methods to enable them to work for problems with multiple, conflicting objectives \cite{roijers2013survey}. RL algorithms generally assume that the environment is a Markov Decision Process (MDP) in which the agent is provided with a scalar reward after each action, and must aim to learn the policy which maximises the long-term return based on those rewards. In contrast MORL algorithms operate within multiobjective MDPs (MOMDPs), in which the reward terms are vectors, with each element in the vector corresponding to a different objective. This creates a number of new issues to be addressed by the MORL agent. Most notably there may be multiple policies which may be optimal (in terms of Pareto optimality), and which policy the agent should learn is not immediately obvious.

In the utility-based paradigm of MORL \cite{roijers2013survey,zintgraf2015quality} it is assumed that the preferences of the user can be defined in terms of a parameterised utility function $f$, and that the aim of the agent should be to learn the policy which produces vector returns which maximises the utility to the user as defined by $f$.

Various approaches have been explored for the form of the utility function -- some may be better suited to express the preference of the user within a particular problem domain, while others offer benefits from an algorithmic perspective. A simple weighted linear scalarisation has been widely used because of its simplicity (for example,  \cite{barrett2008learning,castelletti2010tree, perez2009responsive}). Linear scalarisation transforms an MOMDP into an equivalent single-objective MDP, and enables existing RL approaches to be directly applied \cite{roijers2013survey}. However for many tasks this may not be able to accurately represent the utility of the user, and so may fail to discover the policy which is optimal with regards to their true utility. As a result numerous non-linear scalarisation functions have been explored in the literature (for example,  \cite{gabor1998multi,van2013scalarized, van2013hypervolume}) -- these tend to produce algorithmic complications, but also may better represent the true preferences of the user.

As well as the choice of scalarisation function and  parameters, a second factor must be considered within this utility-based paradigm -- the time-frame over which the utility is being maximised. \citet{roijers2013survey} identified two distinct possibilities. The agent may aim to maximise the expected scalarised return (ESR). That is, it is assumed the returns are first scalarised, and then this agent aims for the policy which maximises the expected value of that scalar. This ESR approach is suited to problems where the aim is to maximise the expected outcome within any individual episode. For example, when producing a treatment plan for a patient which trades off the likelihood of a cure versus the extent of negative side-effects - any individual patient will only undergo this treatment once, and so they care about the utility obtained within that specific episode.

In other contexts we may be concerned about the mean utility received over multiple episodes. In this situation the agent should aim to maximise the scalarised expected return (SER) - that is, it estimates the expected vector return per episode, and then maximises the scalarisation of that expected return. As demonstrated in \citet{roijers2018multi}, the optimal policy for a particular MOMDP under the ESR and SER setting may differ considerably, even if the same scalarisation function and parameters are used in both cases.

As noted by \citet{roijers2018multi} and \citet{ruadulescu2019equilibria} much of the existing work in MORL has considered SER optimization, although this has often been implicit rather than explicitly stated. Much of this SER-focused work has been based on benchmark environments such as those of \citet{vamplew2011empirical}, the majority of which are deterministic MOMDPs.

In this paper we demonstrate by example that the model-free value-based methods previously widely used in MORL research may fail to maximise the SER utility when applied to MOMDPs with stochastic state transitions.

\section{Space Traders: An Example Stochastic MOMDP}
\begin{figure*}
\centering
\includegraphics[width=\textwidth]{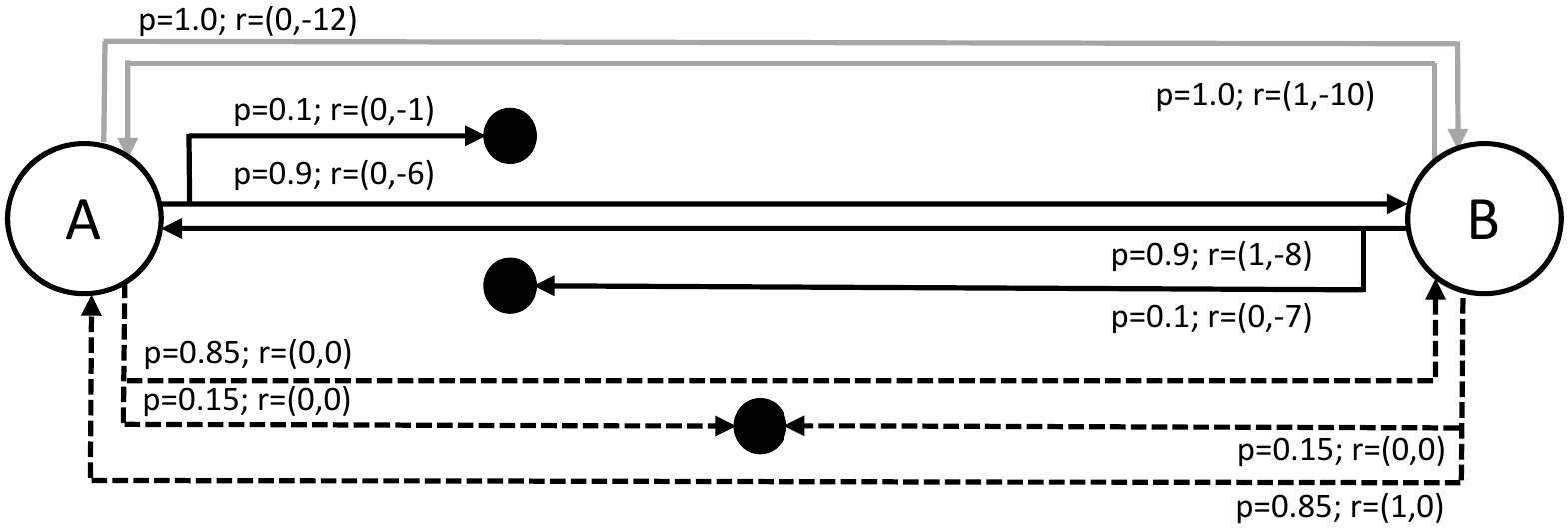}
\caption{\label{fig:space-traders} The Space Traders MOMDP. Solid black lines show the Direct actions, solid grey lines show the Indirect actions, and dashed lines indicate Teleport actions. Sold black circles indicate terminal (failure) states.}
\end{figure*}

As shown in Figure \ref{fig:space-traders} the Space Traders MOMDP is a finite-horizon task with a horizon of 2 time-steps. It consists of two non-terminal states, with three actions available in each state. The agent starts at its home planet (state A) and must travel to another planet (state B) to deliver a shipment, and then return to State A with the payment. The agent receives a reward with two elements - the first is 0 on all actions, except that a reward of 1 is received when the agent successfully returns to state A, while the second element is a negative value reflecting the time taken to execute the action.

There are three possible pathways between the two planets. The direct path (actions shown by solid black lines in Figure \ref{fig:space-traders}) is fairly short, but there is a risk of the agent being waylaid by space pirates and failing to complete the task. The indirect path (grey lines) avoids the pirates and so always leads to successful completion of the mission, but takes longer. Finally the recently developed teleportation system (dashed lines) allows instantaneous transportation, but has a higher risk of failure. The figure also details the probability of success, and the reward for the mission-success and time objectives for each action -- due to variations in local conditions such as solar winds and the location of the space pirates, the time values for the outward and return journeys on a particular path may vary.

Table \ref{tab:action-values} summarises the transition probabilities and rewards of the MOMDP, and also shows the mean immediate reward for each action from each state, weighted by the probability of success.

% \usepackage{multirow}
\begin{table}[]
\begin{tabular}{|l|l|l|l|l|l|}
\hline
State              & Action   & P(success) & \begin{tabular}[c]{@{}l@{}}Reward \\ on\\ success\end{tabular} & \begin{tabular}[c]{@{}l@{}}Reward \\ on\\ failure\end{tabular} & \begin{tabular}[c]{@{}l@{}}Mean\\ reward\end{tabular} \\ \hline
\multirow{3}{*}{A} & Indirect & 1.0        & (0,-12)                                                        & n/a                                                            & (0,-12)                                               \\ \cline{2-6} 
                   & Direct   & 0.9        & (0, -6)                                                        & (0, -1)                                                        & (0, -5.5)                                             \\ \cline{2-6} 
                   & Teleport & 0.85       & (0,0)                                                          & (0,0)                                                          & (0, 0)                                                \\ \hline
\multirow{3}{*}{B} & Indirect & 1.0        & (1, -10)                                                       & n/a                                                            & (1, -10)                                              \\ \cline{2-6} 
                   & Direct   & 0.9        & (1, -8)                                                        & (0, -7)                                                        & (0.9, -7.9)                                           \\ \cline{2-6} 
                   & Teleport & 0.85       & (1, 0)                                                         & (0, 0)                                                         & (0.85, 0)                                             \\ \hline
\end{tabular}
\caption{\label{tab:action-values} The probability of success and reward values for each state-action pair in the Space Traders MOMDP.}
\end{table}

\begin{figure}
\centering
\includegraphics[width=\columnwidth]{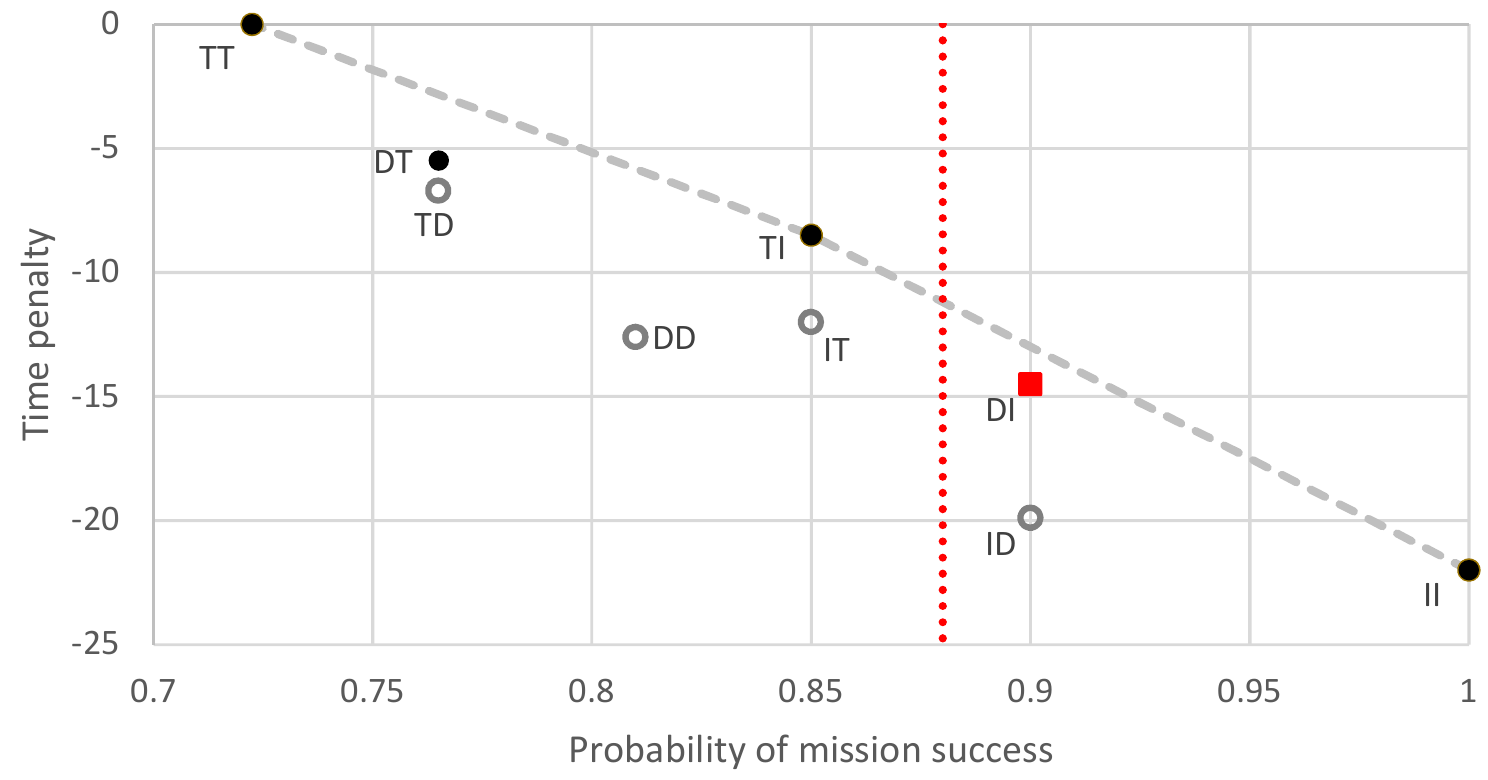}
\caption{\label{fig:policy-values} The mean return per episode for the nine possible deterministic policies for the Space Traders MOMDP. Each policy's return is labelled with a bigram specifying its actions. I, D, T refer to the indirect, direct and teleport actions so, for example, policy DI selects the direct action in state A and the indirect action in state B. Solid markers indicate policies which are members of the Pareto-front, and hollow markers indicate dominated policies. The dashed grey lines illustrate the convex hull formed by mixture combinations of the policies which make up the Convex Coverage Set (CCS). The dashed red vertical line indicates the threshold value of 0.88 for the probability of mission success, and the red square marker is the DI policy which is optimal for that setting of the threshold.}
\end{figure}

As there are three actions from each state there are a total of nine deterministic policies available to the agent. The mean reward per episode for each of these policies is shown in Table \ref{tab:policy-values} and illustrated in Figure \ref{fig:policy-values}. The solid points in the figure highlight the policies which belong to the Pareto front, and the dashed grey line indicates the convex hull (only those policies lying on the convex hull can be located via methods using linear scalarisation -- this set of policies is referred to as the Convex Coverage Set \cite{roijers2013computing}). 

\begin{table}[]
\begin{tabular}{|l|l|l|l|}
\hline
\begin{tabular}[c]{@{}l@{}}Policy\\ identifier\end{tabular} & \begin{tabular}[c]{@{}l@{}}Action in\\ state A\end{tabular} & \begin{tabular}[c]{@{}l@{}}Action in\\ state B\end{tabular} & Mean return     \\ \hline
II                                                          & Indirect                                                    & Indirect                                                    & (1, -22)        \\ \hline
ID                                                          & Indirect                                                    & Direct                                                      & (0.9, -19.9)    \\ \hline
IT                                                          & Indirect                                                    & Teleport                                                    & (0.85, -12)     \\ \hline
DI                                                          & Direct                                                      & Indirect                                                    & (0.9, -14.5)    \\ \hline
DD                                                          & Direct                                                      & Direct                                                      & (0.81, -12.61)  \\ \hline
DT                                                          & Direct                                                      & Teleport                                                    & (0.765, -5.5)   \\ \hline
TI                                                          & Teleport                                                    & Indirect                                                    & (0.85, -8.5)    \\ \hline
TD                                                          & Teleport                                                    & Direct                                                      & (0.765, -6.715) \\ \hline
TT                                                          & Teleport                                                    & Teleport                                                    & (0.7225, 0)     \\ \hline
\end{tabular}
\caption{\label{tab:policy-values} The mean episodic return vector for each of the nine deterministic policies available for the Space Traders MOMDP.}
\end{table}

For the remainder of the paper we will assume that the agent's aim is to minimise the time taken to complete the delivery and return to A, subject to having at least an 88\% probability of successful completion. That is, the scalarisation function $f(\vec{v}) = v_2 $ if $ v_1>0.88 $ and $-\infty$ otherwise. The optimal policy for this aim is to follow the direct path to B and then the indirect path back to A (policy DI).

\section{Applying Model-Free Value-Based MORL Methods to Space Traders}

In this section we will discuss how some of the value-based MORL methods previously used in the literature would perform on the Space Traders MOMDP. All the methods discussed are assumed to be based on a multiobjective extension of model-free value-based RL algorithms such as Q-Learning or SARSA -- for example see \cite[p.~3668]{van2014multi}. For the purposes of this section we will restrict discussion to single-policy methods in which the scalarisation function $f$ is used to filter the multiple Pareto-optimal policies which may be available so as to obtain a single policy which is optimal with regards to $f$. Multiple-policy MORL methods will be discussed in Section \ref{sec-potential-solutions}.

All methods learn vector-valued estimated Q-values, but differ in terms of the scalarisation or ordering method used to perform action-selection, and the characteristics on which the Q-value and policy are conditioned.

\subsection{Linear scalarisation}

A simple approach to MORL is to apply a linear weighted scalarisation to the elements of the Q-value vector prior to selecting the greedy action. As mentioned earlier, this converts the MOMDP into an equivalent MDP, and so the Q-values and action-selection need only be conditioned on the current state of the MDP. However it is well-known that methods using linear scalarisation are unable to identify solutions which do not lie on the convex hull of the Pareto front \cite{vamplew2008limitations}. Clearly from Figure \ref{fig:policy-values} this is the case for policy DI, and so linear methods will not be able to converge to this policy. This result is not surprising and we mention it here simple for the sake of completeness.

\subsection{Non-linear scalarisation}\label{sec:non-linear}

A variety of non-linear scalarisation methods have been explored in the MORL literature \cite{gabor1998multi,van2013scalarized, van2013hypervolume}. The non-linear nature of the scalarisation function means that the assumption of additivity underlying the Bellman equation no longer applies. In order to deal with this, both the choice of action and the Q-values must be conditioned not only on the current state of the environment, but also on rewards received so far by the agent during this episode \cite{geibel2006reinforcement, roijers2018multi}. That is, if the scalarisation function is $f$ then at time $k$ the agent will select the action $a$ which maximises the value of $f(Q(s_k,a,\sum_{t=1}^{k}{r_t})+\sum_{t=1}^{k}{r_t})$.

For the purposes of the following discussion we will assume that $f$ is the thresholded lexicographic ordering operator (TLO) \cite{gabor1998multi, issabekov2012empirical}, and that a thresholding parameter of 0.88 is applied to the first element of the Q-value vector. The intention here is to maximise the value of the second objective (i.e. minimise time), subject to achieving the threshold level for the first objective. If this operator could be applied directly to the mean returns of each policy from Table \ref{tab:policy-values}, then clearly policy DI would be selected.

However if we consider how the TLO operator selects actions during the execution of a policy, then a different result will emerge. Regardless of the path selected at state A, if state B is successfully reached then a zero reward will have been received by the agent. Therefore the choice of action at state B is independent of the previous action. Looking at the mean action values reported in Table \ref{tab:action-values}, it can be seen that action T will be eliminated as it fails to meet the threshold for the first objective, and that action D will be preferred over I as both meet the threshold, and D has a superior value for the time objective. So it can already be seen that this agent will not converge to the desired policy DI.

Knowing that action D will be selected at state B, we can calculate the Q-values for each action at state A, as shown in Table \ref{tab:state-a-values}. The TLO action selector will eliminate actions D and T from consideration as neither meets the threshold of 0.88 for the probability of success. Action I will be selected giving rise to the overall policy ID. Not only is this not the desired DI policy, but as is evident from Figure \ref{fig:policy-values} its average outcome is in fact Pareto-dominated by DI.

\begin{table}[]
\begin{tabular}{|l|l|l|}
\hline
Action in state A & Policy & Q(A, a)         \\ \hline
Indirect          & ID     & (0.9, -19.9)    \\ \hline
Direct            & DD     & (0.81, -12.61)  \\ \hline
Teleport          & TD     & (0.765, -6.715) \\ \hline
\end{tabular}
\caption{\label{tab:state-a-values} The Q-values which will be learned for each action in state A, under the assumption that the Direct action will be selected in State B.}
\end{table}

\section{The Interaction of Local Decision-Making and Stochastic State Transitions}

The failure of the non-linear value-based MORL algorithms on the Space Traders MOMDP can be explained by the analysis of stochastic-transition MOMDPs previously carried out by \citet{bryce2007probabilistic} in the context of probabilistic planning. This analysis has been largely overlooked by MORL researchers so far, and so one of the contributions of this paper is to bring this work to the attention of the MORL research community. 

\begin{figure}
\centering
\includegraphics[width=\columnwidth]{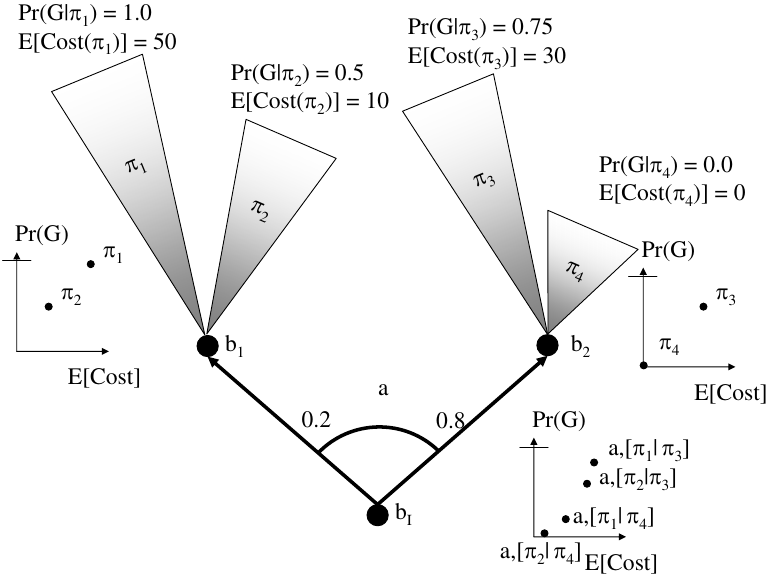}
\includegraphics[width=\columnwidth]{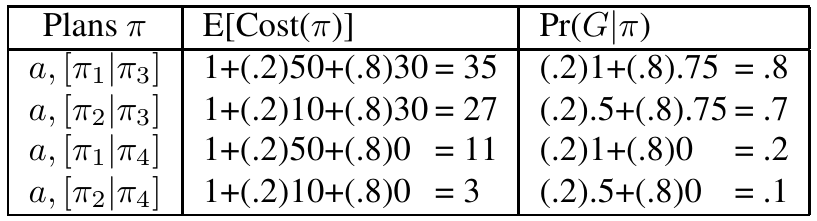}
\caption{\label{fig:bryce} A sample probabilistic planning MOMDP, reproduced from \citet{bryce2007probabilistic}. Executing action a from $b_t$ leads to two branches with probability 0.2 and 0.8. At each of these branches a choice between two sub-plans with different payoffs exists. The aim for the planner is to identify the correct sub-plan to execute at each branch, so as to minimise cost while ensuring successful execution above a fixed probability.}
\end{figure}

Figure \ref{fig:bryce} illustrates a simple MDP reproduced from \citet{bryce2007probabilistic}, with a stochastic branch occurring on the transition from the initial state. The table in the lower half of this figure specifies the mean return for the four possible deterministic policies. Keeping in mind that this MOMDP is phrased in terms of minising cost (rather than maximising the inverse of the cost), it can be seen that unlike Space Traders, there are no Pareto-dominated policies for this MOMDP.\footnote{While clearly illustrating the problem, this MOMDP also lacks the narrative drama of Space Traders!}. 

The aim of the agent is to minimise the cost, subject to satisfying at least a 0.6 probability of success. Within an ESR formulation of the problem (i.e. ensure the probability of success threshold is achieved in each episode), the optimal policy is to select sub-plan $\pi_1$ at branch $b_1$ and $\pi_3$ at branch $b_2$ as both of these sub-plans individually satisfy the probability threshold. However if considered from the SER perspective, the optimal plan is to execute $\pi_2$ at branch $b_1$ and $\pi_3$ at branch $b_2$ -- while $\pi_2$ itself fails to achieve the probability threshold, this branch is executed with a low probability and so the mean outcome of the two sub-plans will achieve the threshold while also producing a significant cost saving.

As identified by \cite{bryce2007probabilistic}, whether the overall policy meets the constraints depends on the probability with which each branch is executed as well as the mean outcome of each branch. Determining the correct sub-plan to follow at each branch requires consideration of the sub-plan options available at each other branch in combination with the probability of branch execution.

This requirement is fundamentally incompatible with the localised decision-making at the heart of model-free value-based RL methods like Q-learning, where it is assumed that the correct choice of action can be determined purely based on information available to the agent at the current state. The provision of additional information such as the sum of rewards received so far in the episode as discussed in Section \ref{sec:non-linear} is insufficient, as it still only provides information about the branch which has been followed in this episode, rather than all possible branches which might have been executed.

The conclusion to be drawn from both this example and Space Traders is that value-based model-free MORL methods are inherently limited when applied in the context of SER optimisation of non-linear utility on MOMDPs with non-deterministic state transitions. These methods may fail to discover the policy which maximises the SER (i.e. the mean utility over multiple episodes). To the best of our knowledge this limitation has not previously been identified in the MORL literature. It is particularly important as the combination of SER, stochastic state transitions and non-linear utility may well arise in important areas of application such as AI safety \cite{vamplew2018human}.

\section{Potential Solutions}\label{sec-potential-solutions}

In this section we will briefly review  and critique various options which may address the issue identified above.

\subsection{ESR Optimisation}

As noted earlier the issue described arises due to the fact that an agent aiming to find a policy optimal with regards to SER must take into account the value which will be received on average by its policy across multiple episodes. Framing the problem in terms of ESR optimisation would eliminate this issue. However ESR is clearly inappropriate for the context of the Space Traders MOMDP. The agent will aim to ensure every episode meets the threshold for the mission-success objective. This can only be achieved by following the strictly safe II policy, which produces results which are far worse for the user's true utility than the DI policy.

\subsection{Non-stationary or non-deterministic policies}\label{sec:non-stat-or-non-det}

\begin{figure}
\centering
\includegraphics[width=\columnwidth]{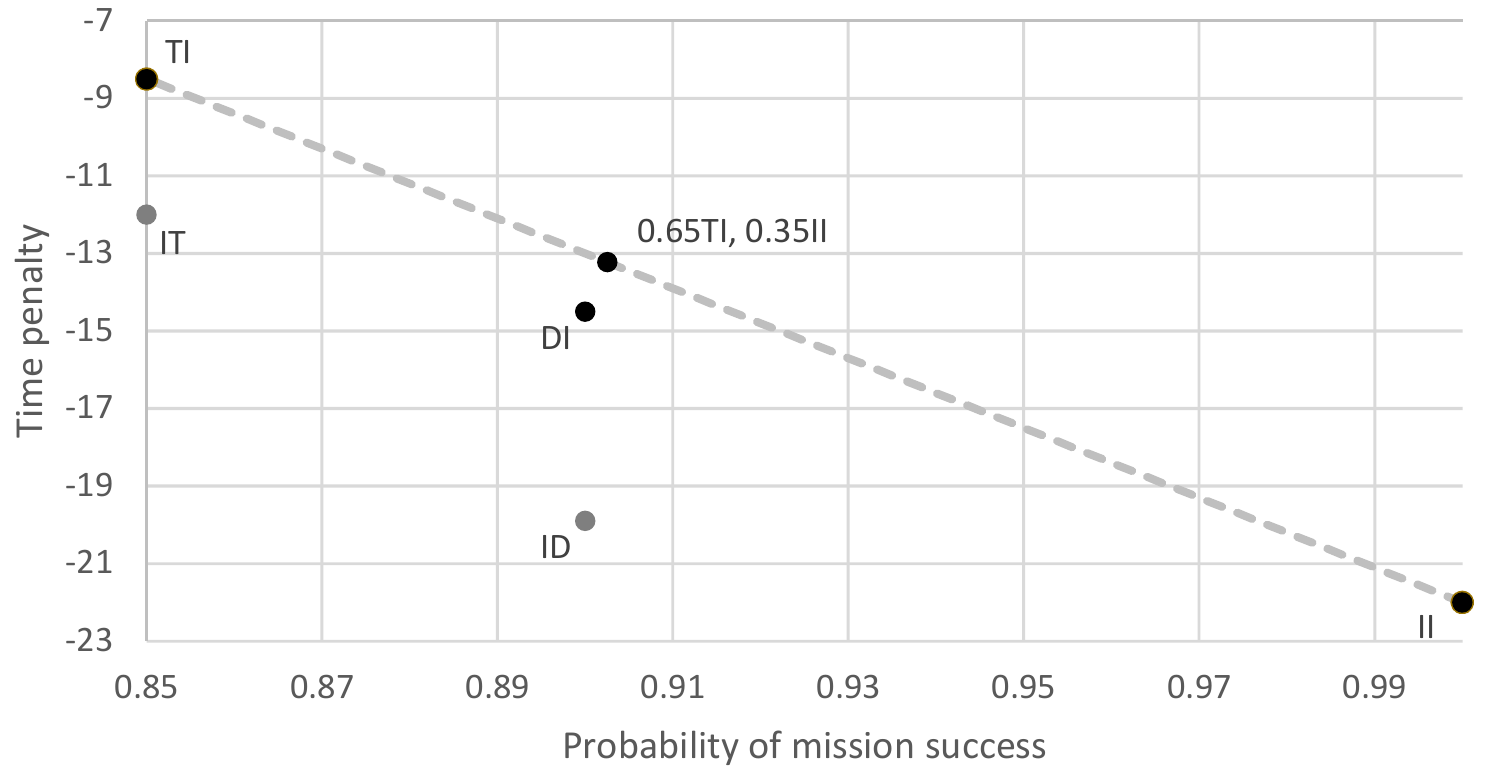}
\caption{\label{fig:space-traders-mixture-policy} The mean return per episode for a mixture policy formed by selecting between the deterministic policies TI and II with probability 0.65 and 0.35 respectively Pareto-dominates the mean return of deterministic policy DI.}
\end{figure}

Previous work has demonstrated that for the SER formulation, or for non-episodic tasks, policies formed from a non-stationary or non-deterministic mixture of deterministic policies can Pareto-dominate deterministic policies \cite{vamplew2009constructing,vamplew2017steering}. For example, a mixture which randomly selects between policies TI and II with appropriate probabilities at the start of each episode can produce a mean outcome which exceeds that of policy DI, as shown in Figure \ref{fig:space-traders-mixture-policy} -- the mixture policy which selects TI with probability 0.65 and II with probability 0.35 achieves a mean return of (0.9025, -13.225) which is superior to the deterministic DI policy with regards to both objectives.

However the use of policies which vary so widely may not be appropriate in all contexts -- for many problems the more consistent outcome produced by a deterministic policy may be preferable, and so methods to find SER-optimal deterministic policies for stochastic MOMDPs are still required.

\subsection{Multi-policy value-based MORL}
As well as the single-policy value-based MORL methods examined in this paper, several authors have proposed multi-policy methods. These operate by retaining multiple value vectors at each state. These can correspond to either all Pareto-optimal values obtainable from that state, or (for purposes of efficiency) be constrained to store only those values which can help construct the optimal value function under some assumptions about the nature of the overall utility function $f$ \cite{roijers2013computing}. Multi-policy algorithms were first proposed for variants of dynamic programming \cite{white1982multi, wiering2007computing} and more recently have been extended to MORL \cite{van2014multi,ruiz2017temporal}.

By propagating back the coverage set of values available at each successor state, these algorithms would correctly identify all potentially optimal policies available at the starting state, and the optimal policy could then be selected at that point -- in the context of Space Traders this would allow for the desired DI policy to be selected. However two issues still need to be addressed. One is ensuring that the agent has a means of determining which action should be performed in each encountered state to align with
the initial choice of policy. Existing algorithms do not necessarily provide
such a means in the context of stochastic transitions. Second, the existing multi-policy MORL algorithms do not have an obvious extension to complex state-spaces where tabular methods are infeasible. Conventional function-approximation methods can not be applied, as the cardinality of the vectors to be stored can vary between states. \citet{vamplew2018non} provides preliminary work addressing this problem, but further work is still required to make this approach practical.  

\subsection{Model-based methods}

As well as describing the difficulties faced by probabilistic planning, \citet{bryce2007probabilistic} also propose a search algorithm known as Multiobjective Looping AO* (MOLAO*) to solve such tasks. As a planning method, this assumes an MOMDP with known state transition probabilities and a finite and tractable number of discrete states. It may be possible to extend this approach by integrating it within model-based RL algorithms which can learn to estimate the transition probabilities and to generalise across states. We are not aware of any prior work which has attempted to do so. However the model-based MORL approach proposed in \citet{wiering2014model} may provide a suitable basis for implementing a reinforcement learning equivalent of MOLAO*.

\subsection{Policy-search methods}
An alternative to value-based approaches is to use policy-search approaches to RL. As these directly maximise the policy as a whole as defined by a set of policy parameters, they do not have the local decision-making issue faced by model-free value-based methods. 

Multiple researchers have proposed and evaluated policy-search methods for multiobjective problems \cite{shelton2001importance, uchibe2007constrained, pirotta2015multi, parisi2017manifold}. One issue to be addressed however is that these methods most naturally produce stochastic policies and as such may have the same problems as faced by the mixture or non-stationary approaches discussed in Section \ref{sec:non-stat-or-non-det}, unless they are modified or constrained so as to ensure convergence to a deterministic policy.

\section{Conclusion}
We have described a stochastic MOMDP and utility function which, despite their seeming simplicity, are not amenable to solution by the widely-used model-free value-based approaches to MORL. While this issue with MOMDPs with stochastic state transitions has previously been described in the context of probabilistic planning \cite{bryce2007probabilistic}, this is the first work to identify the implications for MORL. Our example also demonstrates that under stochastic state-transitions, it is in fact possible for such MORL methods to converge to a Pareto-dominated policy.

The combination of SER optimisation, stochastic state transitions and the need for a deterministic policy are likely to arise in a range of applications (particularly in risk-aware agents), and so awareness of the limitations of some MORL methods to work under these characteristics is important in order to avoid the use of inappropriate methods.

%%%%%%%%%%%%%%%%%%%%%%%%%%%%%%%%%%%%%%%%%%%%%%%%%%%%%%%%%%%%%%%%%%%%%%%%%%%%%%%%%%%%%%%%%%%%%%%%%%%%%%%%%
%% bibliography: see CFP for number of permitted pages

\bibliographystyle{ACM-Reference-Format}  % do not change this line!
\bibliography{sample-bibliography}  % put name of your .bib file here


\begin{thebibliography}{00}

%%% the Web address.

\ifx \showCODEN    \undefined \def \showCODEN     #1{\unskip}     \fi
\ifx \showDOI      \undefined \def \showDOI       #1{#1}\fi
\ifx \showISBNx    \undefined \def \showISBNx     #1{\unskip}     \fi
\ifx \showISBNxiii \undefined \def \showISBNxiii  #1{\unskip}     \fi
\ifx \showISSN     \undefined \def \showISSN      #1{\unskip}     \fi
\ifx \showLCCN     \undefined \def \showLCCN      #1{\unskip}     \fi
\ifx \shownote     \undefined \def \shownote      #1{#1}          \fi
\ifx \showarticletitle \undefined \def \showarticletitle #1{#1}   \fi
\ifx \showURL      \undefined \def \showURL       {\relax}        \fi
% invisible to TeX
\providecommand\bibfield[2]{#2}
\providecommand\bibinfo[2]{#2}
\providecommand\natexlab[1]{#1}
\providecommand\showeprint[2][]{arXiv:#2}

\bibitem[\protect\citeauthoryear{Barrett and Narayanan}{Barrett and
  Narayanan}{2008}]%
        {barrett2008learning}
\bibfield{author}{\bibinfo{person}{Leon Barrett} {and} \bibinfo{person}{Srini
  Narayanan}.} \bibinfo{year}{2008}\natexlab{}.
\newblock \showarticletitle{Learning all optimal policies with multiple
  criteria}. In \bibinfo{booktitle}{{\em Proceedings of the 25th international
  conference on Machine learning}}. \bibinfo{pages}{41--47}.
\newblock


\bibitem[\protect\citeauthoryear{Bryce, Cushing, and Kambhampati}{Bryce
  et~al\mbox{.}}{2007}]%
        {bryce2007probabilistic}
\bibfield{author}{\bibinfo{person}{Daniel Bryce}, \bibinfo{person}{William
  Cushing}, {and} \bibinfo{person}{Subbarao Kambhampati}.}
  \bibinfo{year}{2007}\natexlab{}.
\newblock \showarticletitle{Probabilistic planning is multi-objective}.
\newblock \bibinfo{journal}{{\em Arizona State University, Tech. Rep.
  ASU-CSE-07-006\/}} (\bibinfo{year}{2007}).
\newblock


\bibitem[\protect\citeauthoryear{Castelletti, Galelli, Restelli, and
  Soncini-Sessa}{Castelletti et~al\mbox{.}}{2010}]%
        {castelletti2010tree}
\bibfield{author}{\bibinfo{person}{A Castelletti}, \bibinfo{person}{Stefano
  Galelli}, \bibinfo{person}{Marcello Restelli}, {and} \bibinfo{person}{Rodolfo
  Soncini-Sessa}.} \bibinfo{year}{2010}\natexlab{}.
\newblock \showarticletitle{Tree-based reinforcement learning for optimal water
  reservoir operation}.
\newblock \bibinfo{journal}{{\em Water Resources Research\/}}
  \bibinfo{volume}{46}, \bibinfo{number}{9} (\bibinfo{year}{2010}).
\newblock


\bibitem[\protect\citeauthoryear{G{\'a}bor, Kalm{\'a}r, and
  Szepesv{\'a}ri}{G{\'a}bor et~al\mbox{.}}{1998}]%
        {gabor1998multi}
\bibfield{author}{\bibinfo{person}{Zolt{\'a}n G{\'a}bor},
  \bibinfo{person}{Zsolt Kalm{\'a}r}, {and} \bibinfo{person}{Csaba
  Szepesv{\'a}ri}.} \bibinfo{year}{1998}\natexlab{}.
\newblock \showarticletitle{Multi-criteria reinforcement learning.}. In
  \bibinfo{booktitle}{{\em ICML}}, Vol.~\bibinfo{volume}{98}.
  \bibinfo{pages}{197--205}.
\newblock


\bibitem[\protect\citeauthoryear{Geibel}{Geibel}{2006}]%
        {geibel2006reinforcement}
\bibfield{author}{\bibinfo{person}{Peter Geibel}.}
  \bibinfo{year}{2006}\natexlab{}.
\newblock \showarticletitle{Reinforcement learning for MDPs with constraints}.
  In \bibinfo{booktitle}{{\em European Conference on Machine Learning}}.
  Springer, \bibinfo{pages}{646--653}.
\newblock


\bibitem[\protect\citeauthoryear{Issabekov and Vamplew}{Issabekov and
  Vamplew}{2012}]%
        {issabekov2012empirical}
\bibfield{author}{\bibinfo{person}{Rustam Issabekov} {and}
  \bibinfo{person}{Peter Vamplew}.} \bibinfo{year}{2012}\natexlab{}.
\newblock \showarticletitle{An empirical comparison of two common
  multiobjective reinforcement learning algorithms}. In
  \bibinfo{booktitle}{{\em Australasian Joint Conference on Artificial
  Intelligence}}. Springer, \bibinfo{pages}{626--636}.
\newblock


\bibitem[\protect\citeauthoryear{Parisi, Pirotta, and Peters}{Parisi
  et~al\mbox{.}}{2017}]%
        {parisi2017manifold}
\bibfield{author}{\bibinfo{person}{Simone Parisi}, \bibinfo{person}{Matteo
  Pirotta}, {and} \bibinfo{person}{Jan Peters}.}
  \bibinfo{year}{2017}\natexlab{}.
\newblock \showarticletitle{Manifold-based multi-objective policy search with
  sample reuse}.
\newblock \bibinfo{journal}{{\em Neurocomputing\/}}  \bibinfo{volume}{263}
  (\bibinfo{year}{2017}), \bibinfo{pages}{3--14}.
\newblock


\bibitem[\protect\citeauthoryear{Perez, Germain-Renaud, K{\'e}gl, and
  Loomis}{Perez et~al\mbox{.}}{2009}]%
        {perez2009responsive}
\bibfield{author}{\bibinfo{person}{Julien Perez}, \bibinfo{person}{C{\'e}cile
  Germain-Renaud}, \bibinfo{person}{Bal{\'a}zs K{\'e}gl}, {and}
  \bibinfo{person}{Charles Loomis}.} \bibinfo{year}{2009}\natexlab{}.
\newblock \showarticletitle{Responsive elastic computing}. In
  \bibinfo{booktitle}{{\em Proceedings of the 6th international conference
  industry session on Grids meets autonomic computing}}.
  \bibinfo{pages}{55--64}.
\newblock


\bibitem[\protect\citeauthoryear{Pirotta, Parisi, and Restelli}{Pirotta
  et~al\mbox{.}}{2015}]%
        {pirotta2015multi}
\bibfield{author}{\bibinfo{person}{Matteo Pirotta}, \bibinfo{person}{Simone
  Parisi}, {and} \bibinfo{person}{Marcello Restelli}.}
  \bibinfo{year}{2015}\natexlab{}.
\newblock \showarticletitle{Multi-objective reinforcement learning with
  continuous {P}areto frontier approximation}. In \bibinfo{booktitle}{{\em
  Twenty-Ninth AAAI Conference on Artificial Intelligence}}.
\newblock


\bibitem[\protect\citeauthoryear{R{\u{a}}dulescu, Mannion, Roijers, and
  Now{\'e}}{R{\u{a}}dulescu et~al\mbox{.}}{2019}]%
        {ruadulescu2019equilibria}
\bibfield{author}{\bibinfo{person}{Roxana R{\u{a}}dulescu},
  \bibinfo{person}{Patrick Mannion}, \bibinfo{person}{Diederik~M Roijers},
  {and} \bibinfo{person}{Ann Now{\'e}}.} \bibinfo{year}{2019}\natexlab{}.
\newblock \showarticletitle{Equilibria in multi-objective games: A
  utility-based perspective}. In \bibinfo{booktitle}{{\em Proceedings of the
  adaptive and learning agents workshop (ALA-19) at AAMAS}}.
\newblock


\bibitem[\protect\citeauthoryear{Roijers, Steckelmacher, and Now{\'e}}{Roijers
  et~al\mbox{.}}{2018}]%
        {roijers2018multi}
\bibfield{author}{\bibinfo{person}{Diederik~M Roijers}, \bibinfo{person}{Denis
  Steckelmacher}, {and} \bibinfo{person}{Ann Now{\'e}}.}
  \bibinfo{year}{2018}\natexlab{}.
\newblock \showarticletitle{Multi-objective reinforcement learning for the
  expected utility of the return}. In \bibinfo{booktitle}{{\em Adaptive
  Learning Agents (ALA) workshop at AAMAS}}, Vol.~\bibinfo{volume}{18}.
\newblock


\bibitem[\protect\citeauthoryear{Roijers, Vamplew, Whiteson, and
  Dazeley}{Roijers et~al\mbox{.}}{2013a}]%
        {roijers2013survey}
\bibfield{author}{\bibinfo{person}{Diederik~M Roijers}, \bibinfo{person}{Peter
  Vamplew}, \bibinfo{person}{Shimon Whiteson}, {and} \bibinfo{person}{Richard
  Dazeley}.} \bibinfo{year}{2013}\natexlab{a}.
\newblock \showarticletitle{A survey of multi-objective sequential
  decision-making}.
\newblock \bibinfo{journal}{{\em Journal of Artificial Intelligence
  Research\/}}  \bibinfo{volume}{48} (\bibinfo{year}{2013}),
  \bibinfo{pages}{67--113}.
\newblock


\bibitem[\protect\citeauthoryear{Roijers, Whiteson, and Oliehoek}{Roijers
  et~al\mbox{.}}{2013b}]%
        {roijers2013computing}
\bibfield{author}{\bibinfo{person}{Diederik~M Roijers}, \bibinfo{person}{Shimon
  Whiteson}, {and} \bibinfo{person}{Frans~A Oliehoek}.}
  \bibinfo{year}{2013}\natexlab{b}.
\newblock \showarticletitle{Computing convex coverage sets for multi-objective
  coordination graphs}. In \bibinfo{booktitle}{{\em International Conference on
  Algorithmic DecisionTheory}}. Springer, \bibinfo{pages}{309--323}.
\newblock


\bibitem[\protect\citeauthoryear{Ruiz-Montiel, Mandow, and P{\'e}rez-de-la
  Cruz}{Ruiz-Montiel et~al\mbox{.}}{2017}]%
        {ruiz2017temporal}
\bibfield{author}{\bibinfo{person}{Manuela Ruiz-Montiel},
  \bibinfo{person}{Lawrence Mandow}, {and} \bibinfo{person}{Jos{\'e}-Luis
  P{\'e}rez-de-la Cruz}.} \bibinfo{year}{2017}\natexlab{}.
\newblock \showarticletitle{A temporal difference method for multi-objective
  reinforcement learning}.
\newblock \bibinfo{journal}{{\em Neurocomputing\/}}  \bibinfo{volume}{263}
  (\bibinfo{year}{2017}), \bibinfo{pages}{15--25}.
\newblock


\bibitem[\protect\citeauthoryear{Shelton}{Shelton}{2001}]%
        {shelton2001importance}
\bibfield{author}{\bibinfo{person}{Christian~Robert Shelton}.}
  \bibinfo{year}{2001}\natexlab{}.
\newblock \showarticletitle{Importance sampling for reinforcement learning with
  multiple objectives}.
\newblock  (\bibinfo{year}{2001}).
\newblock


\bibitem[\protect\citeauthoryear{Uchibe and Doya}{Uchibe and Doya}{2007}]%
        {uchibe2007constrained}
\bibfield{author}{\bibinfo{person}{Eiji Uchibe} {and} \bibinfo{person}{Kenji
  Doya}.} \bibinfo{year}{2007}\natexlab{}.
\newblock \showarticletitle{Constrained reinforcement learning from intrinsic
  and extrinsic rewards}. In \bibinfo{booktitle}{{\em 2007 IEEE 6th
  International Conference on Development and Learning}}. IEEE,
  \bibinfo{pages}{163--168}.
\newblock


\bibitem[\protect\citeauthoryear{Vamplew, Dazeley, Barker, and Kelarev}{Vamplew
  et~al\mbox{.}}{2009}]%
        {vamplew2009constructing}
\bibfield{author}{\bibinfo{person}{Peter Vamplew}, \bibinfo{person}{Richard
  Dazeley}, \bibinfo{person}{Ewan Barker}, {and} \bibinfo{person}{Andrei
  Kelarev}.} \bibinfo{year}{2009}\natexlab{}.
\newblock \showarticletitle{Constructing stochastic mixture policies for
  episodic multiobjective reinforcement learning tasks}. In
  \bibinfo{booktitle}{{\em Australasian joint conference on artificial
  intelligence}}. Springer, \bibinfo{pages}{340--349}.
\newblock


\bibitem[\protect\citeauthoryear{Vamplew, Dazeley, Berry, Issabekov, and
  Dekker}{Vamplew et~al\mbox{.}}{2011}]%
        {vamplew2011empirical}
\bibfield{author}{\bibinfo{person}{Peter Vamplew}, \bibinfo{person}{Richard
  Dazeley}, \bibinfo{person}{Adam Berry}, \bibinfo{person}{Rustam Issabekov},
  {and} \bibinfo{person}{Evan Dekker}.} \bibinfo{year}{2011}\natexlab{}.
\newblock \showarticletitle{Empirical evaluation methods for multiobjective
  reinforcement learning algorithms}.
\newblock \bibinfo{journal}{{\em Machine learning\/}} \bibinfo{volume}{84},
  \bibinfo{number}{1-2} (\bibinfo{year}{2011}), \bibinfo{pages}{51--80}.
\newblock


\bibitem[\protect\citeauthoryear{Vamplew, Dazeley, Foale, and
  Choudhury}{Vamplew et~al\mbox{.}}{2018a}]%
        {vamplew2018non}
\bibfield{author}{\bibinfo{person}{Peter Vamplew}, \bibinfo{person}{Richard
  Dazeley}, \bibinfo{person}{Cameron Foale}, {and} \bibinfo{person}{Tanveer
  Choudhury}.} \bibinfo{year}{2018}\natexlab{a}.
\newblock \showarticletitle{Non-functional regression: A new challenge for
  neural networks}.
\newblock \bibinfo{journal}{{\em Neurocomputing\/}}  \bibinfo{volume}{314}
  (\bibinfo{year}{2018}), \bibinfo{pages}{326--335}.
\newblock


\bibitem[\protect\citeauthoryear{Vamplew, Dazeley, Foale, Firmin, and
  Mummery}{Vamplew et~al\mbox{.}}{2018b}]%
        {vamplew2018human}
\bibfield{author}{\bibinfo{person}{Peter Vamplew}, \bibinfo{person}{Richard
  Dazeley}, \bibinfo{person}{Cameron Foale}, \bibinfo{person}{Sally Firmin},
  {and} \bibinfo{person}{Jane Mummery}.} \bibinfo{year}{2018}\natexlab{b}.
\newblock \showarticletitle{Human-aligned artificial intelligence is a
  multiobjective problem}.
\newblock \bibinfo{journal}{{\em Ethics and Information Technology\/}}
  \bibinfo{volume}{20}, \bibinfo{number}{1} (\bibinfo{year}{2018}),
  \bibinfo{pages}{27--40}.
\newblock


\bibitem[\protect\citeauthoryear{Vamplew, Issabekov, Dazeley, Foale, Berry,
  Moore, and Creighton}{Vamplew et~al\mbox{.}}{2017}]%
        {vamplew2017steering}
\bibfield{author}{\bibinfo{person}{Peter Vamplew}, \bibinfo{person}{Rustam
  Issabekov}, \bibinfo{person}{Richard Dazeley}, \bibinfo{person}{Cameron
  Foale}, \bibinfo{person}{Adam Berry}, \bibinfo{person}{Tim Moore}, {and}
  \bibinfo{person}{Douglas Creighton}.} \bibinfo{year}{2017}\natexlab{}.
\newblock \showarticletitle{Steering approaches to Pareto-optimal
  multiobjective reinforcement learning}.
\newblock \bibinfo{journal}{{\em Neurocomputing\/}}  \bibinfo{volume}{263}
  (\bibinfo{year}{2017}), \bibinfo{pages}{26--38}.
\newblock


\bibitem[\protect\citeauthoryear{Vamplew, Yearwood, Dazeley, and Berry}{Vamplew
  et~al\mbox{.}}{2008}]%
        {vamplew2008limitations}
\bibfield{author}{\bibinfo{person}{Peter Vamplew}, \bibinfo{person}{John
  Yearwood}, \bibinfo{person}{Richard Dazeley}, {and} \bibinfo{person}{Adam
  Berry}.} \bibinfo{year}{2008}\natexlab{}.
\newblock \showarticletitle{On the limitations of scalarisation for
  multi-objective reinforcement learning of pareto fronts}. In
  \bibinfo{booktitle}{{\em Australasian Joint Conference on Artificial
  Intelligence}}. Springer, \bibinfo{pages}{372--378}.
\newblock


\bibitem[\protect\citeauthoryear{Van~Moffaert, Drugan, and
  Now{\'e}}{Van~Moffaert et~al\mbox{.}}{2013a}]%
        {van2013hypervolume}
\bibfield{author}{\bibinfo{person}{Kristof Van~Moffaert},
  \bibinfo{person}{Madalina~M Drugan}, {and} \bibinfo{person}{Ann Now{\'e}}.}
  \bibinfo{year}{2013}\natexlab{a}.
\newblock \showarticletitle{Hypervolume-based multi-objective reinforcement
  learning}. In \bibinfo{booktitle}{{\em International Conference on
  Evolutionary Multi-Criterion Optimization}}. Springer,
  \bibinfo{pages}{352--366}.
\newblock


\bibitem[\protect\citeauthoryear{Van~Moffaert, Drugan, and
  Now{\'e}}{Van~Moffaert et~al\mbox{.}}{2013b}]%
        {van2013scalarized}
\bibfield{author}{\bibinfo{person}{Kristof Van~Moffaert},
  \bibinfo{person}{Madalina~M Drugan}, {and} \bibinfo{person}{Ann Now{\'e}}.}
  \bibinfo{year}{2013}\natexlab{b}.
\newblock \showarticletitle{Scalarized multi-objective reinforcement learning:
  Novel design techniques}. In \bibinfo{booktitle}{{\em 2013 IEEE Symposium on
  Adaptive Dynamic Programming and Reinforcement Learning (ADPRL)}}. IEEE,
  \bibinfo{pages}{191--199}.
\newblock


\bibitem[\protect\citeauthoryear{Van~Moffaert and Now{\'e}}{Van~Moffaert and
  Now{\'e}}{2014}]%
        {van2014multi}
\bibfield{author}{\bibinfo{person}{Kristof Van~Moffaert} {and}
  \bibinfo{person}{Ann Now{\'e}}.} \bibinfo{year}{2014}\natexlab{}.
\newblock \showarticletitle{Multi-objective reinforcement learning using sets
  of pareto dominating policies}.
\newblock \bibinfo{journal}{{\em The Journal of Machine Learning Research\/}}
  \bibinfo{volume}{15}, \bibinfo{number}{1} (\bibinfo{year}{2014}),
  \bibinfo{pages}{3483--3512}.
\newblock


\bibitem[\protect\citeauthoryear{White}{White}{1982}]%
        {white1982multi}
\bibfield{author}{\bibinfo{person}{DJ White}.} \bibinfo{year}{1982}\natexlab{}.
\newblock \showarticletitle{Multi-objective infinite-horizon discounted Markov
  decision processes}.
\newblock \bibinfo{journal}{{\em Journal of mathematical analysis and
  applications\/}} \bibinfo{volume}{89}, \bibinfo{number}{2}
  (\bibinfo{year}{1982}), \bibinfo{pages}{639--647}.
\newblock


\bibitem[\protect\citeauthoryear{Wiering and De~Jong}{Wiering and
  De~Jong}{2007}]%
        {wiering2007computing}
\bibfield{author}{\bibinfo{person}{Marco~A Wiering} {and}
  \bibinfo{person}{Edwin~D De~Jong}.} \bibinfo{year}{2007}\natexlab{}.
\newblock \showarticletitle{Computing optimal stationary policies for
  multi-objective markov decision processes}. In \bibinfo{booktitle}{{\em 2007
  IEEE International Symposium on Approximate Dynamic Programming and
  Reinforcement Learning}}. IEEE, \bibinfo{pages}{158--165}.
\newblock


\bibitem[\protect\citeauthoryear{Wiering, Withagen, and Drugan}{Wiering
  et~al\mbox{.}}{2014}]%
        {wiering2014model}
\bibfield{author}{\bibinfo{person}{Marco~A Wiering}, \bibinfo{person}{Maikel
  Withagen}, {and} \bibinfo{person}{M{\u{a}}d{\u{a}}lina~M Drugan}.}
  \bibinfo{year}{2014}\natexlab{}.
\newblock \showarticletitle{Model-based multi-objective reinforcement
  learning}. In \bibinfo{booktitle}{{\em 2014 IEEE Symposium on Adaptive
  Dynamic Programming and Reinforcement Learning (ADPRL)}}. IEEE,
  \bibinfo{pages}{1--6}.
\newblock


\bibitem[\protect\citeauthoryear{Zintgraf, Kanters, Roijers, Oliehoek, and
  Beau}{Zintgraf et~al\mbox{.}}{2015}]%
        {zintgraf2015quality}
\bibfield{author}{\bibinfo{person}{Luisa~M Zintgraf}, \bibinfo{person}{Timon~V
  Kanters}, \bibinfo{person}{Diederik~M Roijers}, \bibinfo{person}{Frans
  Oliehoek}, {and} \bibinfo{person}{Philipp Beau}.}
  \bibinfo{year}{2015}\natexlab{}.
\newblock \showarticletitle{Quality assessment of MORL algorithms: A
  utility-based approach}. In \bibinfo{booktitle}{{\em Benelearn 2015:
  Proceedings of the 24th Annual Machine Learning Conference of Belgium and the
  Netherlands}}.
\newblock


\end{thebibliography}

\end{document}